\documentclass[conference]{IEEEtran}
\IEEEoverridecommandlockouts

\usepackage{cite}
\usepackage{amsmath,amssymb,amsfonts}
\usepackage{algorithmic}
\usepackage{graphicx}
\usepackage{textcomp}
\usepackage{xcolor}
\usepackage{multirow}
\usepackage{hyperref}
\usepackage{enumerate}

\usepackage{booktabs}

\usepackage{enumitem}  

\hypersetup{hidelinks}
\def\BibTeX{{\rm B\kern-.05em{\sc i\kern-.025em b}\kern-.08em
    T\kern-.1667em\lower.7ex\hbox{E}\kern-.125emX}}
\begin{document}

\title{InceptionCaps: A Performant Glaucoma Classification Model for Data-scarce Environment\\
}

\author{\IEEEauthorblockN{Gyanendar Manohar}
\IEEEauthorblockA{\textit{Department of Computer Science} \\
\textit{Munster Technological University}\\
Cork, Ireland \\
0000-0001-7289-0688}
\and
\IEEEauthorblockN{Ruairi O'Reilly}
\IEEEauthorblockA{\textit{Department of Computer Science} \\
\textit{Munster Technological University}\\
Cork, Ireland \\
0000-0001-7990-3461}
}

\maketitle

\begin{abstract}
Glaucoma is an irreversible ocular disease and is the second leading cause of visual disability worldwide. Slow vision loss and the asymptomatic nature of the disease make its diagnosis challenging. Early detection is crucial for preventing irreversible blindness. Ophthalmologists primarily use retinal fundus images as a non-invasive screening method. Convolutional neural networks (CNN) have demonstrated high accuracy in the classification of medical images. Nevertheless, CNN's translation-invariant nature and inability to handle the part-whole relationship between objects make its direct application unsuitable for glaucomatous fundus image classification, as it requires a large number of labelled images for training. 
This work reviews existing state of the art models and proposes InceptionCaps, a novel capsule network (CapsNet) based deep learning model having pre-trained InceptionV3 as its convolution base, for automatic glaucoma classification.
InceptionCaps achieved an accuracy of 0.956, specificity of 0.96, and AUC of 0.9556, which surpasses several state-of-the-art deep learning model performances on the RIM-ONE v2 dataset. The obtained result demonstrates the robustness of the proposed deep learning model.

\end{abstract}


\section{Introduction}
\IEEEPARstart{G}laucoma is a leading cause of visual impairment in the world \cite{b1}.
It is an ocular disease caused by damage to the optic nerve, which sends images to the brain. The health of the optic nerve is essential for good vision, and it can be damaged by high intraocular pressure in the eyes, causing irreversible vision loss. The majority of glaucoma patients are asymptomatic, with symptoms only appearing as the disease progresses to the advanced stage. It is common in adults in their 60s and above. Additionally, ethnic origin, genetics, and related medical problems such as diabetes, high blood pressure, and eye injury/surgery increase the risk of glaucoma.

Glaucoma is mainly related to the retinal and optic nerve fiber layer \cite{b2}, optic nerve assessment is essential for disease detection. It also brings different types of abnormality in the optic disc (OD) region \cite{b3}.
Hence the OD region is the central region of interest (ROI) in digital fundus images for glaucoma detection. 
The number of glaucoma patients is rising and is projected to reach 95 million by 2030 and 11.8 million by 2040 \cite{b4}. However, due to the complex and time-consuming testing process and a lack of expert ophthalmologists, the world population will suffer badly from preventable blindness. A computer-aided system for the automatic detection of glaucoma using retinal fundus images would aid ophthalmologists in making a diagnosis, reduce human errors and shorten the individual's examination time.

This work is motivated to develop a performant deep learning-based neural network for glaucoma classification that can be trained in a data-scarce environment.
Four state-of-the-art (SOTA) models, both CNN and CapsNet-based, for glaucoma classification are reproduced.
The main contribution of this work is to demonstrate the utility of transfer learning (TL) over data augmentation in limited training dataset scenarios. Consequently InceptionCaps, a novel deep learning (DL) model consisting of a pre-trained InceptionV3 and capsule network for glaucoma classification is proposed.

Artificial neural networks (ANN) have gained notable success in classification and object detection with advancements in computer vision and machine learning (ML). Over the past decade, extensive research has been conducted for computer-aided detection of glaucoma from retinal fundus images using imagery transformation, supervised machine learning, semi-supervised machine learning, and DL techniques. 

CNN are a type of deep-feed forward artificial neural network used across the domain for image classification. The main benefit of CNN architectures is their ability to extract highly discriminating features at multiple levels of abstraction.
The main challenges are as follow:

\begin{enumerate}
    \item[i)] CNNs can predict an object's presence in an image but have no knowledge about the position of one object with respect to another. So, when an object is moved from its position, the CNN cannot detect this movement. This is the translation-invariant property of CNN.
    \item[ii)] CNNs requires a large training dataset. Data is required from all viewpoints to generalize the result. The pooling layer of CNN discards lots of features value from the feature map and requires significant quantities of training data to compensate for the loss. The large training dataset requirement makes the training of CNN slow and resource-intensive.
\end{enumerate}   
Retinal fundus imagery has many differentiating features present in the lower-level pixel. CNN's pooling layer discarding part of them makes it difficult for a model to train on a small dataset. The ethical and data protection issues associated with patient fundus imagery makes it challenging to acquire data in sufficient quantities for training CNNs. Table \ref{tab:dataset} denotes publicly available glaucomatous fundus image datasets used for the automatic classification of glaucoma. The main challenges with the referenced  datasets are as follows:

\begin{enumerate}
        \item Datasets have fewer training images, which is unsuitable for training a CNN-based neural network. 
       \item Images in the dataset are not diverse. For example, it does not include images taken from different camera models and patients from different ethnicities.       
       \item Datasets have a significant class imbalance issue.
\end{enumerate}

CapsNet \cite{b7} overcame some of the shortcomings of CNNs. Its main building blocks are capsule, a small group of neurons where each neuron represents the various properties of a particular part of the image and outputs a vector. The magnitude of the vector determines the likelihood of the presence of a particular object or region in the image represented by the capsule. Furthermore, the capsule network strives for viewpoint equivariance, i.e., any change in viewpoint at input results in a similar variance in the output’s viewpoint.
If an image is divided into different parts, each part will have its characteristic, which a set of characteristic features can represent. A neuron can represent a specific property of an object or region within the picture. So, a group of neurons can represent several properties of the object. 
CapsNet requires relatively fewer images for training \cite{b8} and performs robustly with class imbalance issues \cite{b9}, which makes it an appealing choice for medical image classification.

\begin{table}[tb]
\centering
\caption{Public Retinal fundus image dataset for glaucoma classification}
\begin{tabular}{lllll}
      \hline
      \textbf{Dataset~\textsubscript{Ref}}&  \textbf{No. Images}&
      \textbf{Glau}& \textbf{Normal}& \textbf{Task} \\
      \hline
      RIM-ONE v1~\textsubscript{\cite{b10}} & 131& 39& 92&  CLS,SEG  \\
      RIM-ONE v2~\textsubscript{\cite{b10}}& 455& 200& 255& CLS  \\
      RIM-ONE v3~\textsubscript{\cite{b10}}& 159& 74&85&CLS,SEG  \\
      RIM-ONE dl~\textsubscript{\cite{b11}}& 485& 172&313&CLS  \\          
      DRISHTI-GS~\textsubscript{\cite{b17}}& 101& 70&31&CLS  \\
      ACRIMA~\textsubscript{\cite{b27}}& 705& 396&309&CLS  \\
      ODIR~\textsubscript{\cite{b28}}& 5000&207 &1135&CLS  \\
      G1020~\textsubscript{\cite{b29}}& 1020& 296&724& CLS,SEG \\      
     \hline      
\end{tabular}

\vspace{.20cm}

\textbf{*CLS} is Classification \textbf{*SEG} is Segmentation \textbf{*Glau} is Glaucomatous
\textbf{*No.Images} is Number of Images
\label{tab:dataset}
\end{table}

Transfer learning (TL) and data augmentation are widely used due to data scarcity in the medical domain. Various data augmentation strategies have been successfully applied in the medical domain to force the model to learn discriminative image features and have improved model performance. Image transformations made to an image during augmentation may cause the generated image to lose its discriminative features, degrading model performance. The suitability of a combination of augmentation techniques for different medical images is under-researched. 

DL-based generative adversarial networks (GAN) have also been used to produce synthetic retinal images for training CNN models for eye diseases classification \cite{b33}. However, GAN-based models are complex and require high-end computational resources, which can limit their utility \cite{b35}. Hence, this work focuses on TL, and DL model architecture for glaucoma classification from fundus images. 

The reminder of the paper is organised as follows: Section \ref{relatedwork} provides an overview of related work in glaucoma classification from fundus imagery using CapsNet and TL-based DL model. Section \ref{Methodology} details the dataset selection,  design of the baseline DL model, and CapsNet. Section \ref{resultsanddiscussion} presents results and discusses the findings. Section \ref{conclusion} details the conclusion of this work and outlines suggested future work.

\section{Related Work} \label{relatedwork}

A significant body of research exists on the automatic classification of retinal diseases, including glaucoma, from the fundus imagery using DL. 
The automated feature extraction capabilities of DL make it a preferred choice for fundus image segmentation and classification. CNNs are widely utilised for this image classification task. However, it is difficult to train CNNs from scratch for glaucoma classifications due to the need for large training datasets, which are scarce. Consequently, researchers have used TL and multiple data augmentation methods when training glaucoma classification models. 

 \begin{table*}[t]
 \caption{DL based glaucoma classification model with their architecture, dataset used, augmentation techniques, Optimiser,  performance metrics and limitation}
	\centering
 \begin{tabular}{p{1.85cm}p{2.9cm}p{0.3cm}p{1.0cm}p{1.8cm}p{0.8cm}p{0.5cm}p{0.5cm}p{0.5cm}p{0.5cm}p{0.3cm}p{1.8cm}}
      \toprule
      \textbf{Model~\textsubscript{Ref}}& \textbf{ Architecture}&  \textbf{TLA}&\textbf{Dataset} & \textbf{Aug}& \textbf{Opt} & \textbf{AUC} & \textbf{SEN} &  \textbf{SPE} & \textbf{ACC} &  \textbf{Re-Imp}& \textbf{Limitations} \\
      \midrule
      Bander~\textsubscript{\cite{b25}}& 
      AlexNet, SVM & FE&
       R1 v1 & -
      & -
      &
      - & 
       0.85 & 
       0.908 &
      0.882 & 
      \checkmark&
      -\\
      Cerentini~\textsubscript{\cite{b26}}&
      GoogLeNet &FT&
      R1 v1, v2, v3& 
      Rot, re-scaling, Gv, Gna 
      &
      ADAM
      &
      0.862&-&-&-& \checkmark&-\\

       Batista~\textsubscript{\cite{b11}}
       &V16, V19, RNet, DNet, MobileNet, InV3&FT &R1 dl &
      Hvf,Rot, zoom&RMSProp
      &&&&& \checkmark&-\\

      Maheshwari~\textsubscript{\cite{b30}}& 
      AlexNet& FT&
      R1 v2 &
      Lbp
      &SGDM&
      - & 
       1 & 
      0.975 &
      0.989 &&
      Custom prep\\
      
      Sallam~\textsubscript{\cite{b31}}& 
      Pre-trained CNN & FT&
       LAG & -
      &&
      0.869&  
      0.869 & 
      - &
      - &&
      Private dataset \\

      Lima~\textsubscript{\cite{b32}}& ResNet,& FE &R1 v2&-&&0.957 &-&-&0.901&&Custom prep\\
      &INet, LR&& R1 v3&&& 0.860&-&-&0.805&&\\

      Phankokkruad~\textsubscript{\cite{b34}}& V16, RNetV2, Xpt, InV3&FT & REFUGE& -& & 0.9719&-&-&-&&Dataset unavailable\\

    \midrule
    \multicolumn{9}{c}{\textbf{Capsule Network based models for glaucoma classification}}\\
    \midrule

      dos Santos~\textsubscript{\cite{b20}}& CapsNet &-& R1 v2&-&RMSProp&-&-&-&0.909& \checkmark&\\
      Gaddipatis~\textsubscript{\cite{b19}}&CapsNet& -&-&-&-&-&-&-&&&3D OCT Images\\ 
      
      \bottomrule
 \end{tabular}
\vspace{.10cm}

	*\textbf{TLA} is Transfer Learning Approach
    *\textbf{Aug} is Augmentation
    *\textbf{Opt} is Optimiser
    *\textbf{AUC} is Area Under the ROC Curve
    *\textbf{SEN} is sensitivity
	*\textbf{SPE} is Specificity    
    *\textbf{ACC} is accuracy
    *\textbf{Re-Imp} is Re-Implemented
    *\textbf{Prep} is preprocessing    
    *\textbf{FE} is Feature Extraction
    *\textbf{FT} is Fine Tuning
    *\textbf{R1} is RIM-ONE
    *\textbf{Rot} is Random rotation
    *\textbf{Lbp} is Local binary pattern
    *\textbf{Gv} is Gamma variation
    *\textbf{Gna} is Gaussian noise addition
    *\textbf{Hvf} is Horizontal and vertical flip
    *\textbf{V} is VGG
    *\textbf{RNet} is ResNet50
    *\textbf{DNet} is DenseNet121
    *\textbf{Xpt} is Xception
    *\textbf{InV3} is InceptionV3
    *\textbf{INet} is InceptionNet
    \label{tab:transferlearningbasedglaucomaclassification}
\end{table*}

\subsection{Transfer Learning }
A TL-based approach reduces model training time and can improve classification performance. There are two approaches to TL, namely feature extraction and fine-tuning, both of which have been used for developing glaucoma classification models (See Table \ref{tab:transferlearningbasedglaucomaclassification}). 

\textbf{Feature extraction}: CNNs have excellent feature extraction capability. Pre-trained CNNs extract features from the training input image, which acts as input to the linear classifiers like RF, LR, and SVM to do the actual classification \cite{b25}. \textbf{Fine tuning}: The pre-trained CNNs contain many layers. The initial layers learn low-level features of the image, whereas higher layers learn the more significant structures. For using it with a different dataset, the pre-trained network's top layer is removed, and a new fully connected layer is added. The resulting network is trained on a new dataset after freezing the weight of the pre-trained network \cite{b11} \cite{b26}.

\subsection{Capsule Network} \label{capsnetrb}

As denoted in Table \ref{tab:transferlearningbasedglaucomaclassification}, CapsNet have been used for retinal disease classification. CapsNet~\cite{b7}, when originally proposed, used routing algorithms for information propagation across the layer. The routing process is computationally expensive and consequently slows down the training process. Furthermore, using the same size convolution kernel for feature generation is insufficient to capture sufficient feature depth from an image where the target objects have complex representation \cite{b22}. Several improvements, such as multi-scale convolution layer \cite{b23} and modified routing algorithm \cite{b24}, have been introduced to overcome these limitations.

\section{Methodology} \label{Methodology}

Four SOTA models, are re-implemented as indicated in Table \ref{tab:transferlearningbasedglaucomaclassification}. Each SOTA model undergoes more extensive experimentation in order to evaluate the models' performance. The following stages of a DL workflow for classifying glaucomatous fundus images are detailed: i) Data selection, ii) Data preprocessing, iii) Model preparation/selection, iv) Training, validating, and tuning the model, and v) Class prediction for test data. 

\noindent 
\begin{itemize}[left=4pt]
    \item[i)] Dataset selection: Retinal Image database for Optic Nerve Evaluation (RIM-ONE) \cite{b10}  is a highly cited publicly available dataset. RIM-ONE contains four versions v1, v2, v3, and dl. RIM-ONE v2 and RIM-ONE dl \cite{b11} contain fundus images segmented by experts. 
    ACRIMA \cite{b27} is a glaucoma labeled public dataset from the ACRIMA project. This dataset is well suited for classification as images are segmented around the optic disc region.
    \item[ii)]  Pre-processing: All images are resized as per selected pre-trained CNN model input requirement, and their pixels' values are normalized in the range of [0,1]. 
    \item[iii)] Model Selection: Results of all SOTA models could not be faithfully reproduced due to limitations such as the use of a private dataset, custom preprocessing steps, or omission of details relating to model architecture. As denoted in Table \ref{tab:transferlearningbasedglaucomaclassification}, TL-based models that were trained on publicly available data, without any specialized preprocessing, were selected for the purposes of this work.
    \item[iv)] Training and Validation: The training and validation methods used are adapted from their descriptions. Most hyper-parameter values used were taken from the SOTA model selected for reproduction, and further experimentation is highlighted in the relevant section. Missing hyperparameter values were empirically determined by observing the model performance. It is observed that during the training of a model, the epoch point where validation loss is minimum does not match with point where validation accuracy is maximum. The difference is in the range of 1-2\%. As validation loss is the more stable parameter for comparing the performance, this work utilises the model checkpointing method to select the model where validation loss is minimum.
    \item[v)] Evaluation Metrics: The selection of evaluation metrics is important as model performance and their interpretation can vary. Accuracy is the most commonly used evaluation metric for glaucoma classification, followed by Area Under the Receiver Operating Characteristics Curve (AUC), precision, and recall. 
    \item[] \textbf{Note}: All works used Tensor flow and Keras, OpenCV, Matplotlib and Matlab live script for the model implementation on Google Collaborator with high-speed RAM.
\end{itemize}

\subsection{DL model details and their reproduction steps} \label{transferlearningbased}

\subsubsection{AlexNet based model}
\label{alexnetbasedmodel}

Al-Bander \cite{b25} model was re-implemented using MATLAB as Keras library does not provide pre-trained AlexNet.

\subsubsection{GoogLeNet based model}\label{gnetbasedmodel}
For reproducing Cerentini \cite{b26} model, the top layer of an Imagenet trained InceptionV3 is removed, and a new fully-connected layer with two class output softmax added. The pre-trained weights of InceptionV3 are kept frozen and the model is trained with a batch size of 32, and a learning rate of 0.0001. The validation accuracy is monitored with a patience value of 10. 

\subsubsection{Transfer learning based RIM-ONE DL validation}\label{Rimonedlvalidation}
The model detailed in RIM-ONE dl's original paper \cite{b11} is implemented. 
Random rotation of (-30,30) degrees, vertical and horizontal flips, and zoom ranges 0.8 to 1.2 are applied to the input image for data augmentation. The pre-trained models VGG16, VGG19, ResNet50, DenseNet121, and MobileNetV2 trained on the Imagenet dataset for TL are used. 

The fully connected layers of the pre-trained model are replaced with new fully connected layers for the two-class output softmax layer.	In addition, the GlobalAveragePooling2D layer was added to the convolutional base. The resultant model is trained with a batch size of 32. The model is trained and fine-tuned by firstly, freezing the pre-trained layer's weight and training the newly added top layer with the learning rate 2e-5, secondly, unfreezing the pre-trained layers and jointly training the complete network with the learning rate 1e-5. The training uses 5-fold cross-validation by keeping a .8:.2 ratio of training data for the training and validation set.

The resultant model is trained again on a full training dataset using the steps used previously with cross-validation. Finally, the resultant model classification performance is evaluated with the test dataset. 

The RIM-ONE dl dataset has a class imbalance issue, so accuracy is not chosen as the primary metric for measuring the model performance. This is to avoid the biases exhibited by the model. Instead, AUC-ROC was used as a reference for evaluating network performance. This metrics is complemented with sensitivity and specificity. 

\subsubsection{Capsule network based model}
\label{capsnetimp}

The CapsNet-based model in \cite{b10} was selected for reproduction and establishing baseline performance for glaucoma classification. The original work used RIM-ONE v2 \cite{b10} and Drishti-GS \cite{b17} dataset for model performance evaluation. 

CapsNet tries to model every part of the image, which increases the computational complexity, slows down the training, and reduces the accuracy. As part of preprocessing, images are resized to 64$\times$64 pixels and, histogram equalization (HE) is applied to each image to enhance the contrast, structural and visual aspects before passing to the network. This makes the model robust against the sharp variation in the image contrast, which can impact the image features learned by the network. 

The model has two convolutional layers followed by a fully connected layer. The first layer has 256 filters with a 9$\times$9 kernel with stride value one and ReLu activation function. The second layer is the primary capsule layer, having eight-dimensional 32 convolutional capsules with 9$\times$9 cores and a stride value of 2. The third layer is the digitcap layer, which produces 16-dimensional vector output for each class. This work focuses on the full-color space only. The final model is trained for 200 epochs with a learning rate of 0.0001.

\subsection{Experimental Analysis} \label{experimentation}
In order to evaluate the SOTA DL models presented in Section \ref{transferlearningbased}, a comparative analysis is performed. In this context, each model will adopt data preprocessing, augmentation, TL, and classification as reported originally in the first instance. The following experiments are to be carried out for each model:

\begin{enumerate}   
\item Evaluate model performance on RIM-ONE v2, RIM-ONE dl, and ACRIMA dataset.
\item Study the model performance with and without data augmentation.
\item Evaluate model performance for different test train split ratios. 
\item Solidify the usability of TL techniques under limited training data availability by using different CNNs.
\item Evaluate model generalisation capability across different datasets.
\end{enumerate}

For the CapsNet model, the first layer is a convolutional layer that generates the feature maps from the input image. It has been discovered during the related work review in Section \ref{capsnetrb} that a single convolutional layer may not always be able to extract sufficient features to get decent performance. This makes it imperative to study the CaspNet classification performance with changes in the initial convolutional layer. For the CapsNet, the following experimentation is undertaken:

\begin{enumerate}
    \item Change filter count and the kernel size of the convolutional layer of the discussed CapsNet architecture in the previous section.
    \item Use the TL approach with CapsNet and study the impact on classification performance of the resulting model. 
\end{enumerate}
    
The intent is to utilize the automatic feature extraction power of pre-trained CNNs whose top layers are replaced with the primary capsule and the class capsule with two outputs. 
The pre-trained part of the model extracts high-level feature details, and the capsule part extracts low-level features from the fundus image. The resulting model is trained end to end for glaucoma classification.

\section{Results and Discussion} 
\label{resultsanddiscussion}
This section presents a quantitative evaluation of model performance for glaucoma classification. Each model's performance is first compared to the stated result in the model's original research article. Then, the outcome of all of the model's experimentation as described in Section \ref{experimentation} is presented and discussed with the underlying model to maintain flow coherency.

\subsection{AlexNet based model \textbf{MIV-A}}\label{AlexNetBasedCNN}

The AlexNet-based CNN described in Section \ref{alexnetbasedmodel} is hereafter referred to as M\ref{AlexNetBasedCNN}. 
Table \ref{tab:baseline1} denotes the classification performance of M\ref{AlexNetBasedCNN} (AlexNet, SVM) and the alternate pre-trained CNN, classifier permutations. 
M\ref{AlexNetBasedCNN}'s performance is comparable to that reported in \cite{b25}, confirming the model's reproducibility. As M\ref{AlexNetBasedCNN} performed well on the RIM-ONE v2 dataset without any additional preprocessing or data augmentation, it is important to investigate the model's classification performance across a variety of datasets, the impact of data augmentation, the efficacy of various models, and the utility of a TL-based model in a limited training dataset scenario.

\begin{table}[tb]
\caption{Glaucoma classification performance for different architecture, train test split ratio across different datasets}
\begin{center}
\begin{tabular}{p{1.5cm}p{1.0cm}p{0.6cm}lp{0.6cm}p{0.6cm}p{0.6cm}}
\hline
\textbf{Model}& \textbf{Dataset}&\textbf{Tr:Te}& \textbf{Aug}&\textbf{ACC}& \textbf{SEN}& \textbf{SPE}  \\
\hline
      Bander~\textsubscript{\cite{b25}}& R1 v2&0.7:0.3&No & 0.882 & 0.85  & 0.908 \\      

 M\ref{AlexNetBasedCNN} &  R1 v2&0.7:0.3&No&0.889 & 
      0.883 & 0.894 \\          

  M\ref{AlexNetBasedCNN} & R1 dl &0.7:0.3 &No& 0.890 & 0.865 & 0.904 \\

 M\ref{AlexNetBasedCNN} & ACRIMA &0.7:0.3 &No& 0.943 & 0.966 & 0.914 \\

 M\ref{AlexNetBasedCNN} & R1 v2& 0.6:0.4 &No& 0.879 & 0.875 & 0.882\\

 M\ref{AlexNetBasedCNN} & R1 v2&0.8:0.2 &No& 0.901 & 0.850 & 0.941\\

 M\ref{AlexNetBasedCNN} & R1 v2&0.8:0.2 &Rot,T& 0.901 & 0.850 & 0.941\\
 M\ref{AlexNetBasedCNN} & R1 v2&0.8:0.2 &T,Ref& 0.901 & 0.850 & 0.941\\
 M\ref{AlexNetBasedCNN} & R1 v2&0.8:0.2 &T& 0.901 & 0.850 & 0.941\\

MV2, LR & R1 v2 & 0.8:0.2&No&0.934 &     0.931 &     0.936  \\
MV2, SGD & R1 v2 & 0.8:0.2&No&0.905 &     0.948 &     0.873  \\
MV2, SVM & R1 v2 & 0.8:0.2&No&0.875 &     0.862 &     0.886  \\

ResNet50, LR &R1 v2 & 0.8:0.2&No&  0.890 &   0.827 &     0.936  \\

ResNet50, RF & R1 v2 & 0.8:0.2&No& 0.875 &     0.793 &     0.936  \\

VGG19, LR &R1 v2 & 0.8:0.2&No&  0.883  &     0.793 &     0.949  \\

VGG16, LR &R1 v2 & 0.8:0.2&No&  0.875  &     0.810 &     0.924  \\

DN121, SGD &R1 v2 & 0.8:0.2&No&  0.868  &     0.844 &     0.886  \\

DN121, SVM &R1 v2 & 0.8:0.2&No&  0.861  &     0.775 &     0.924  \\

DN121, LR &R1 v2 & 0.8:0.2&No&  0.861  &     0.793 &     0.911  \\

InceptionV3, LR &R1 v2 & 0.8:0.2&No&  0.817 &     0.758 &     0.860  \\

InceptionV3, SGD &R1 v2 & 0.8:0.2&No&  0.810  &     0.689 &     0.898  \\

InceptionV3, SVM &R1 v2 & 0.8:0.2&No&  0.817  &     0.689 &     0.911  \\
 
    \hline
\end{tabular}
\end{center}
\vspace{.20cm}
*\textbf{R1} is RIM-ONE
*\textbf{Tr:Te} is Train:Test
*\textbf{Aug} is Augmentation
*\textbf{Rot} is Random Rotation
*\textbf{T} is Translation
*\textbf{Ref} is Reflection
*\textbf{ACC} is Accuracy
*\textbf{SEN} is Sensitivity
*\textbf{SPE} is Specificity
*\textbf{MV2} is MobileNetV2
*\textbf{DN121} is DenseNet121

\label{tab:baseline1}
\end{table}

\subsubsection{Performance across different datasets}
M\ref{AlexNetBasedCNN}'s classification performance is nearly identical for RIM-ONE dl and RIM-ONE
v2 but achieved higher accuracy for the ACRIMA dataset, which has more images, and the image quality is optimized for deep learning.
\subsubsection{Varying train test split ratio}
For the train test ratio of 0.8:0.2, M\ref{AlexNetBasedCNN} achieved the highest specificity, which
directly measures the true negative value. With the increase in ratio, the model had
more training examples, which increased the model’s sensitivity and reduced specificity.

\subsubsection{Effect of data augmentation} \label{alexnetexpAug}

M\ref{AlexNetBasedCNN} classification performance improved by 1\%, which is not a considerable improvement given TL alone achieved an accuracy of around 88\%. With translation
alone, M\ref{AlexNetBasedCNN} classification performance didn't improve, whereas with random rotation, the model classification performance degraded.
Hence, TL is considered a better approach to augmentation for training the CNN in a limited data environment.

\subsubsection{Pre-trained CNN and classifier permutations}\label{alexnetexp4}

In Table \ref{tab:baseline1}, the combination of MobileNetV2 with LR was more performant than other pre-trained CNNs' on the RIM-ONE v2 dataset. The MobileNetV2 based CNN is hereafter referred to as M\ref{alexnetexp4}.

\subsubsection{Model generalisation on different dataset}
The number of images in RIM-ONE v2 and RIM-ONE dl are limited and do not include images from different ethnicities, different camera models, or the rare cases of glaucoma. M\ref{AlexNetBasedCNN} does not learn sufficient unique patterns to generalize well for different fundus image datasets. 
The result indicates that the model cannot generalize well on diverse test input and should be trained on a larger, more diverse dataset to enable more generalised performance.

\subsection{GoogLeNet based model \textbf{MIV-B}}\label{googlenetcnn}
The GoogleNet based CNN described in Section \ref{gnetbasedmodel} is hereafter referred to as M\ref{googlenetcnn}. 
Table \ref{tab:baseline2} denotes the classification performance of M\ref{googlenetcnn}. M\ref{googlenetcnn}'s performance is comparable to that reported in \cite{b26}, confirming the model's reproducibility.

\begin{table}[tb]
\caption{Glaucoma classification performance with different train test split ration, architecture across datasets}
\centering
  \begin{tabular}{p{0.6cm}p{1.50cm}p{1.2cm}p{0.6cm}ll}
    \toprule
    \textbf{Model}& \textbf{Architecture} & \textbf{Dataset}& \textbf{Tr:Te}& \textbf{Aug} & \textbf{ACC} \\
    \midrule
    \cite{b26} & GoogleNet & R1 v2 & 0.9:0.1 & No &  0.862 \\
    M\ref{googlenetcnn}&GoogleNet & R1 v2 & 0.9:0.1 & No &  0.869 \\
    M\ref{googlenetcnn}&GoogleNet & R1 dl & 0.9:0.1 & No &  0.869 \\
    M\ref{googlenetcnn}&GoogleNet & ACRIMA & 0.9:0.1 & No &  0.901 \\
    M\ref{googlenetcnn}&GoogleNet & R1 v2 & 0.7:0.3 & No &  0.897 \\
    M\ref{googlenetcnn}&GoogleNet & R1 v2 & 0.9:0.1 & Yes &  0.913 \\
    M\ref{googlenetcnn}&GoogleNet & R1 v2 & 0.8:0.2 & No &  0.912 \\
    M\ref{googlenetcnn}&GoogleNet & R1 v2 & 0.7:0.3 & No &  0.897 \\

    M\ref{googlenetcnn}&InceptionV3 & R1 v2 & 0.9:0.1 & No &  0.901 \\
    M\ref{googlenetcnn}&MobileNet & R1 v2 & 0.9:0.1 & No &  0.890 \\
    M\ref{googlenetcnn}&DenseNet121 & R1 v2 & 0.9:0.1 & No &  0.890 \\
    M\ref{googlenetcnn}&MobileNetV2 & R1 v2 & 0.9:0.1 & No &  0.879 \\
    M\ref{googlenetcnn}&Xception & R1 v2 & 0.9:0.1 & No &  0.868 \\
    M\ref{googlenetcnn}&VGG16 & R1 v2 & 0.9:0.1 & No &  0.835 \\
    M\ref{googlenetcnn}&IRV2 & R1 v2 & 0.9:0.1 & No &  0.824 \\
    M\ref{googlenetcnn}&VGG19 & R1 v2 & 0.9:0.1 & No &  0.802 \\
    M\ref{googlenetcnn}&ResNet50 & R1 v2 & 0.9:0.1 & No &  0.681 \\
    
    \bottomrule
  \end{tabular}  
  \vspace{.25cm}
  
  *\textbf{Tr:Te} is Train:Test
  *\textbf{Aug} is Augmentation
  *\textbf{R1} is RIM-ONE
  *\textbf{ACC} is Accuracy
  *\textbf{IRV2} is InceptionResNetV2
  \label{tab:baseline2}
\end{table}

\subsubsection{Performance across different datasets}

M\ref{googlenetcnn}'s classification performance is nearly identical for RIM-ONE dl and RIM-ONE
v2 but achieved higher accuracy for the ACRIMA dataset, which has more images, and the images quality is optimized for the deep learning model.

\subsubsection{Varying train test split ratio}
 M\ref{googlenetcnn} classification performance improved as the number of training images increased. This is because the model learns more fundus image attributes as more training data is provided. 

\subsubsection{Effect of data augmentation} \label{googlenetaug}

M\ref{googlenetcnn} classification accuracy improved by 4\%, when the horizontal and vertical flip, random rotation of 10 degrees, and zooming were applied during the model fine-tuning.

\subsubsection{Pre-trained CNNs} \label{googlenetcnnexp4}
Models' performance varied by changing the pre-trained CNN. The resultant network consisting of InceptionV3 is more performant than other pre-trained CNNs. The InceptionV3-based CNN is hereafter referred to as M\ref{googlenetcnnexp4}.

\subsubsection{Model generalisation on different datasets}

RIM-ONE dl and RIM-ONE v2 datasets have images from the same source, so model performance is almost at the same level. However, the images of the ACRIMA dataset have a different resolution, and the source and model trained on the RIM-ONE dataset do not perform well. This indicates that M\ref{googlenetcnn} cannot generalize well on different datasets unless trained on larger dataset with diverse images. 

\begin{table}[tb]
\caption{ M\ref{AlexNetBasedCNN} and M\ref{googlenetcnn} generalisation capability across the datasets}
\begin{center}
\begin{tabular}{llllll}
\hline
\textbf{Model}&\textbf{Training Dataset} & \textbf{Test Dataset} & \textbf{Accuracy} \\ 
\hline
       
      M\ref{AlexNetBasedCNN} &RIM-ONE v2 & RIM-ONE dl & 
      0.828 \\

       && ACRIMA & 
      0.495 \\
      \hline      
      M\ref{AlexNetBasedCNN}&RIM-ONE dl & RIM-ONE v2 & 
      0.904 \\
      
       && ACRIMA & 
      0.575 \\
      
      \hline
       
      M\ref{googlenetcnn} & RIM-ONE v2 & RIM-ONE dl & 
      0.178 \\
      
       && ACRIMA & 
      0.436 \\
      \hline
       M\ref{googlenetcnn}& RIM-ONE dl & RIM-ONE v2 & 
      0.195 \\
      
       && ACRIMA & 
      0.521 \\
      
      \hline
      
 \end{tabular}
	\label{tab:baseline2exp5}
\end{center}
\label{tab:baseline2exp3}
\end{table}

\subsection{Transfer learning based RIM-ONE DL validation \textbf{MIV-C}}\label{rimonedlvalidation}
Different CNN-based models are developed as described in Section \ref{Rimonedlvalidation} to validate the reported performance for glaucoma classification on the RIM-ONE dl dataset.
Table \ref{tab:baseline3} denotes the glaucoma classification performance metrics reported in  \cite{b11} and the reproduced metrics using the steps mentioned in Section \ref{Rimonedlvalidation}. 

The TL-based RIM-ONE DL validation is hereafter referred to as M\ref{rimonedlvalidation}. Data augmentation techniques random rotation (-30,30) degree, zoom, vertical, and horizontal flip are used to avoid overfitting during the model training.

M\ref{rimonedlvalidation}'s glaucoma classification performance is approximately 0.07 points less than the reported AUC value in \cite{b11} for most pre-trained networks.
The difference in outcome could be because of the implementation mismatch. Unfortunately, the complete details about the network architecture used are omitted in \cite{b11}.

\begin{table}[tb]
\caption{M\ref{rimonedlvalidation} glaucoma classification performance on RIM-ONE dl dataset with and without augmentation}
	\centering
 \begin{tabular}{lp{0.35cm}p{0.35cm}p{0.45cm}p{0.35cm}p{0.35cm}p{0.45cm}p{0.35cm}p{0.35cm}p{0.45cm}}
      \hline &\multicolumn{3}{c}{\textbf{Reported}} & \multicolumn{3}{c}{\textbf{Achieved - Aug}}& \multicolumn{3}{c}{\textbf{Without Aug}}\\
      \hline
       \textbf{P Model} & \textbf{AUC} & \textbf{SEN} & \textbf{ACC} & \textbf{AUC} & \textbf{SEN} & \textbf{ACC} & \textbf{AUC} & \textbf{SEN} & \textbf{ACC} \\
      
      \hline
      
      VGG16 & 
      0.983 & 
      0.961 &
      0.924 &
      0.917 &
      0.989 &
      0.938
     & 0.897 & 0.957&0.872
      \\
      
      VGG19 & 
      0.986 & 
      1.000&
      0.931&
      0.920&
      0.936&
      0.924
      &0.876&0.872&0.878
      \\
      
      Xception & 
      0.977 & 
      0.980 &
      0.917 &
      0.901 &
      0.957 &
      0.917
      &0.890&0.936&0.871
      \\
      
      ResNet50 & 
      0.975 & 
      0.980 &
       0.911 &
       0.500 & 
       1.000&
       0.643
       &0.643&1.000&0.500
      \\
      
      MNetV2 & 
      0.973 & 
      0.942 &
       0.904&
       0.712&
       0.925&
       0.773
       &0.760&0.861&0.719
      \\
      
      DenseNet & 
      0.972 & 
      0.961 &
       0.904&
       0.938&
       0.914&
       0.931
       &0.890&0.904&0.884
      \\
      
      MNet & 
      0.971 & 
      0.961 &
       0.931&
       0.939&
       0.936&
       0.938
       &0.890&0.904&0.884
      \\
      
      IRV2 & 
      0.968 & 
      0.980 &
       0.911&
       0.895&
       0.925&
       0.904
       &0.842&0.872&0.830
      \\
     
      IV3 & 
      0.959 & 
      0.942 &
       0.890&
       0.863&
       0.957&
       0.890
       &0.849&0.840&0.852
      \\
     
      \hline
 \end{tabular}
 
\vspace{.20cm}
	*\textbf{AUC} is areas under ROC curve
    *\textbf{SEN} is sensitivity
    *\textbf{ACC} is accuracy
    *\textbf{Aug} is Augmentation
    *\textbf{MNet} is MobileNet
    *\textbf{IRV2} is InceptionResNetV2
    *\textbf{IV3} is InceptionV3
    *\textbf{P Model} is Pre-trained Model
    \label{tab:baseline3}
\end{table}

\subsubsection{Impact of data augmentation }\label{rimonedlaug}
    \label{rimonedlexp1}

M\ref{rimonedlvalidation}'s classification performance improved by an average of 5\% with data augmentation.

\subsubsection{Performance across different datasets}\label{rimonedlexp2}
The model has performed well for the ACRIMA dataset. 

\textbf{M\ref{rimonedlexp2}}
The Xception-based model has achieved the highest AUC value of 0.921. Xception based model is hereafter referred to as M\ref{rimonedlexp2}.
\begin{table}[tb]
\caption{M\ref{rimonedlvalidation} glaucoma classification performance on RIM-ONE v2 and ACRIMA dataset without augmentation }
	\centering
 \begin{tabular}{lllllll}
      \hline &\multicolumn{3}{c}{\textbf{ACRIMA}} &\multicolumn{3}{c}{\textbf{RIM-ONE v2}} \\
      \hline
       Pre-trained Model & AUC & SEN & ACC & AUC & SEN & ACC \\
      
      \hline
	  vgg16 & 0.973 & 0.982 & 0.971 &0.906 & 0.853& 0.912 \\
	  vgg19 & 0.950 & 0.948 & 0.950 & 0.899 & 0.978 & 0.901 \\
	  ResNet50 & 0.733 & 0.551 & 0.765 & 0.500 & 1.000 & 0.450 \\
	  DenseNet121 & 1.00 & 1.00 & 1.00 & 0.869 & 0.878 & 0.868 \\
	  MobileNet & 0.979& 0.982& 0.978 & 0.913 & 0.926& 0.912 \\
	  MobileNetV2 &0.586 & 0.172&0.659 & 0.704 & 0.609 & 0.714 \\
	  InceptionV3 &0.985 & 0.982& 0.985 & 0.864 & 0.829 & 0.868 \\
	  IRV2 &0.973 & 0.982& 0.971& 0.884 & 0.829 & 0.890 \\
	  xception & 0.973 & 0.982 & 0.971& 0.921& 0.902 & 0.923 \\
	  
	        \hline
 \end{tabular}	
    \label{tab:baseline3exp2}
\vspace{.25cm}
    
    *\textbf{ACC} is Accuracy
    *\textbf{AUC} is Area Under the ROC Curve
    *\textbf{SEN} is Sensitivity
    *\textbf{IRV2} is InceptionResNetV2
    
\end{table}

Glaucoma classification performance of M\ref{AlexNetBasedCNN}, M\ref{googlenetcnn}, and M\ref{rimonedlvalidation} confirms that TL techniques are helpful for a limited training example scenario. 

\subsection{Capsule network-based model \textbf{MIV-D}} \label{capsnetdlmodel}

Table \ref{tab:capsnetbaseline} denotes the achieved classification performance of CapsNet discussed in Section \ref{capsnetimp}. 
 M\ref{capsnetdlmodel} performance for glaucoma classification
 is comparable to that reported in \cite{b20} for all the evaluation metrics values, confirming the model's reproduciblity. The images from the Drishti-GS1 dataset are full fundus images, and no segmentation was performed on the images before being passed to the network during training. The model performed better than M\ref{AlexNetBasedCNN} and M\ref{googlenetcnn}, even with full fundus images and without data augmentation. The HE has enhanced the image characteristics, enabling improved learning and, hence, improved classification accuracy.

\begin{table}[tb]
\caption{M\ref{capsnetdlmodel} glaucoma classification performance across different dataset with different image size and convolution layer}
	\centering
 \begin{tabular}{p{0.15cm}p{0.65cm}p{1.35cm}p{1.0cm}p{1.3cm}p{0.4cm}p{0.3cm}}
      \hline \textbf{HE} & \textbf{Model} & \textbf{Dataset}& \textbf{Img Size}&\textbf{Con Layer}& \textbf{ACC} &\textbf{AUC} \\
      \hline
       Yes & \cite{b20} &R1 v2, GS1& 64$\times$64&256(9$\times$9) &0.909 & 0.904 \\
      Yes& M\ref{capsnetdlmodel} &R1 v2, GS1& 64$\times$64& 256(9$\times$9) &0.892 &  0.890  \\
      \hline      
      No & \cite{b20} &R1 v2, GS1& 64$\times$64&256(9$\times$9) & 0.882 &  0.891 \\
      No &M\ref{capsnetdlmodel} &R1 v2, GS1& 64*64&256(9$\times$9)  & 0.875 &  0.873 \\
      \hline
      Yes &M\ref{capsnetdlmodel} &R1 v2& 64$\times$64&256(9$\times$9) &  0.912 & 0.913  \\
      No &M\ref{capsnetdlmodel} &R1 v2& 64$\times$64& 256(9$\times$9) & 0.879 & 0.876  \\
      Yes &M\ref{capsnetdlmodel} &R1 v2& 128$\times$128&256(9$\times$9) &   0.857 &  0.852 \\
       Yes &M\ref{capsnetdlmodel} &R1 dl& 64$\times$64& 256(9$\times$9)  &0.904 &  0.873  \\
      No &M\ref{capsnetdlmodel} &R1 dl& 64$\times$64&256(9$\times$9)  & 0.876 &  0.861  \\
      Yes &M\ref{capsnetdlmodel} &ACRIMA& 64$\times$64& 256(9$\times$9) & 0.893 & 0.894  \\
      No &M\ref{capsnetdlmodel} &ACRIMA& 64$\times$64& 256(9$\times$9) & 0.964 & 0.964  \\
      No &M\ref{capsnetkf} &R1 v2& 64$\times$64& 128(9$\times$9) & 0.901 & 0.901  \\
      No &M\ref{capsnetkf} &R1 v2& 64$\times$64& 64(9$\times$9) & 0.890 & 0.889  \\
      No &M\ref{capsnetkf} &R1 v2& 64$\times$64& 64(7$\times$7) & 0.824 &  0.824  \\
      No &M\ref{capsnetkf} &R1 v2& 64$\times$64& 128(9$\times$9), 64(9$\times$9) & 0.891 & 0.901  \\
      No &M\ref{capsnetkf} &R1 v2& 64$\times$64& 32(3$\times$3), 64(5$\times$5), 128(7$\times$7) & 0.912 & 0.909  \\
      No & M\ref{capsnetpretrainedcnnbase} &R1 v2& 64$\times$64&InceptionV3&0.956&0.955\\
      No & M\ref{capsnetpretrainedcnnbase} &R1 v2& 64$\times$64&MobileNet& 0.945& 0.943\\
      No & M\ref{capsnetpretrainedcnnbase} &R1 v2& 64$\times$64&DenseNet121& 0.912 &0.904\\
      No & M\ref{capsnetpretrainedcnnbase} &R1 v2& 64$\times$64&MobileNetV2&0.802 &0.791\\
      No & M\ref{capsnetpretrainedcnnbase} &R1 v2& 64$\times$64&Xception&0.934& 0.929\\
      No & M\ref{capsnetpretrainedcnnbase} &R1 v2& 64$\times$64&VGG16& 0.934 &0.931\\      
      No & M\ref{capsnetpretrainedcnnbase} &R1 v2& 64$\times$64&VGG19& 0.879 &0.876\\      
      \hline
      
 \end{tabular}
\vspace{.25cm}

    *\textbf{HE} is Histogram equalisation
    *\textbf{ACC} is Accuracy
    *\textbf{AUC} is Area Under the ROC Curve
    *\textbf{R1} is RIM-ONE
    *\textbf{GS1} is DRISHTI-GS1
    *\textbf{Con} is Convolutional
	\label{tab:capsnetbaseline}
\end{table}

\subsubsection{Performance for different image sizes}

M\ref{capsnetdlmodel}'s classification performance degraded with large image size, and the reason discussed in Section \ref{capsnetimp}.
Therefore, for all further experimentation, unless stated otherwise, the images are resized to 64$\times$64 pixels.

\subsubsection{Performance across different datasets}\label{capsnetdifferentdataset}
M\ref{capsnetdlmodel} glaucoma classification performance for ACRIMA, RIM-ONE dl and RIM-ONE v2 dataset is better than many of the
TL-based models discussed in previous sections for the RIM-ONE v2 and
the RIM-ONE dl dataset.
Also, HE shouldn't be applied to all types of images as evident from the degradation in model performance for the ACRIMA dataset.

\subsubsection{Impact of different kernels and filters}
\label{capsnetkf}

The CapsNet with different filter counts and kernels is hereafter referred to as M\ref{capsnetkf}. 
Table \ref{tab:capsnetbaseline} denotes M\ref{capsnetkf} glaucoma classification performance on RIM-ONE v2 dataset. 
The performance of the CapsNet model varies with the change in filter count and kernel size of the base convolutional layer. A large filter count with a larger kernel size produces a rich feature map, which enhances the models' classification performance.

The main takeaway of Section \ref{capsnetkf} is that a single convolutional layer with a single kernel size cannot extract sufficient image features. However, a deeper convolutional base increases the training data requirement, which defies the main advantage of capsule network usability with a limited training set. These experiments provided an intuition of InceptionCaps, described in the next section.

\subsubsection{Pre-trained CNN and capsule network}
\label{capsnetpretrainedcnnbase}

Pre-trained CNNs are good feature extractors, as indicated by the performance of M\ref{AlexNetBasedCNN}, and M\ref{googlenetcnn}.
The CNN outputs the spatial representation of the input image instead of the single convolutional layer, as discussed in Section \ref{capsnetimp}.
The CapsNet with CNN as the base is hereafter referred to as M\ref{capsnetpretrainedcnnbase}.
 The model performance is evaluated on the RIM-ONE v2 dataset. Resizing and pixel value rescaling were applied to the input image.

\subsubsection{InceptionCaps}\label{inceptioncaps} The most performant model from the experimental analysis utilised InceptionV3 as the base layer of the CapsNet (See Table IX). It achieved an accuracy of 0.956 and AUC of 0.9556 on the RIM-ONE v2 dataset. Additionally, this outperforms all prior results presented from Section \ref{AlexNetBasedCNN} to Section \ref{rimonedlvalidation} as well as the original SOTA models selected for re-implementation from Table \ref{tab:transferlearningbasedglaucomaclassification}.

In isolation, the convolutional layer would discard low-level features from the input fundus image, critical for glaucoma classification. InceptionCaps utilizes low-level feature representation of CapsNet and feature extraction of CNNs. The boost in the classification performance is due to this unique characteristic of the capsule network.

InceptionV3 was selected as the base of the CapsNet architecture. MobileNet achieved a comparable accuracy (see Table \ref{tab:capsnetbaseline}). The primary difference between these two models is that MobileNet uses depthwise separable convolution, which reduces the number of learnable parameter requirements and slightly degrades performance compared to InceptionV3, which uses standard convolution.

\begin{table}[tb]
\caption{Glaucoma classification performance across DL model implemented and SOTA model on RIM-ONE v2 dataset }
	\centering
 \begin{tabular}{p{1.5cm}p{2.20cm}p{0.4cm}p{0.4cm}p{0.4cm}p{0.4cm}}
      \hline
      Model & Architecture &  ACC & AUC &  SEN & SPE \\
      
      \hline
      \cite{b25} &TL - AlexNet&
      
      0.882 & &0.85 & 0.908\\
      
      \cite{b26}&TL - GoogLeNet&
      
      0.86 &- &- & -\\
      
      \cite{b30} &TL - AlexNet &  \textbf{0.989} &- &- & -\\
      \cite{b32} &TL &  0.901 & \textbf{0.957}& - & -\\
      M\ref{alexnetexp4} &TL - MobileNetV2&   0.934 & 0.933 &     0.931 &     0.936  \\
      
      M\ref{googlenetcnnexp4} &TL - InceptionV3 &  0.901 &   0.896 &     0.853 &         0.94  \\
      
       M\ref{rimonedlexp2}&TL - Xception &   0.921& 0.902 & 0.923 &  \\

      \textbf{InceptionCaps} & \textbf{TL - CapsNet}& 
      \textbf{0.956} &
      \textbf{0.955} &
      \textbf{0.951} & 
      \textbf{0.96}
      
      \\

      \hline
 \end{tabular}
 
\vspace{.25cm}
 *\textbf{TL} is Transfer learning
 *\textbf{ACC} is Accuracy
 *\textbf{SEN} is Sensitivity
 *\textbf{SPE} is Specificity
    
    \label{tab:capsnetTransferlearning}
\end{table}

\subsection{Final Discussion}

Table \ref{tab:capsnetTransferlearning} denotes the glaucoma classification performance on the RIM-ONE v2 dataset of the novel DL model proposed in this work (InceptionCaps), reproduced SOTA models (M\ref{AlexNetBasedCNN}, M\ref{googlenetcnn}, M\ref{rimonedlvalidation}, M\ref{capsnetdlmodel}) and SOTA models that could not be reproduced ~\cite{b30, b32}~(due to the omission of custom image preprocessing techniques and details relating to the model implementation).

InceptionCaps achieved the highest accuracy in all but one instance, in the context of TL-based CNNs M\ref{AlexNetBasedCNN}, M\ref{googlenetcnn} and M\ref{rimonedlvalidation}. The ability to retain low-level features enabled by the CapsNet assisted with the performance gain. CapsNet works well for smaller image sizes, and the performance starts to degrade as the images size increases (see Table \ref{tab:capsnetbaseline}).

CapsNet with a CNN base achieved classification accuracy of 0.956, which outperformed \cite{b11},\cite{b25} and \cite{b26} for glaucoma classification from the fundus images. 
InceptionCaps obtained the highest specificity of 0.96, indicating that most glaucomatous eyes were detected as glaucomatous. In addition, InceptionCaps achieved the highest accuracy without any additional preprocessing and data augmentation. 

Compared to SOTA CNNs' reporting performance on the RIM-ONE v2 dataset, only the model in \cite{b30} performed better than InceptionCaps. In \cite{b30} the training image count is augmented by treating each channel of input RGB image as an unique instance, and an LBP transformation is applied to it. In contrast, no direct or indirect data augmentation was applied during the training of InceptionCaps.

As seen in Section \ref{alexnetexpAug}, \ref{googlenetaug} and \ref{rimonedlaug}, geometric transformation-based data augmentation techniques enhance the model learning capability and improves classification performance. 
Images generated post-augmentation lose unique features essential for the classification task and degrade models' performance, as seen in Table \ref{tab:baseline1}. 

The proposed CapsNet-based model with InceptionV3 as convolution base is a unique, performant and novel approach to glaucoma classification. While attempts have been made with vanilla CapsNet or pre-trained CNN's, but their combination has not been previously attempted.

\section{Conclusion and Future Work}\label{conclusion}
This work reproduces four state-of-the-art deep learning models for glaucoma classification from fundus imagery. The utility of TL over data augmentation under a limited training dataset scenario is demonstrated. The efficacy of CapsNet for glaucoma classification is evaluated and the impact of classification performance with changes in the convolutional layers and routing iteration is investigated. 

InceptionCaps, a novel DL model consisting of a InceptionV3 and capsule network for glaucoma classification, is presented robustly and transparently, detailing all its artifacts and model details. The model achieved an accuracy of 0.956 and an AUC of 0.9556, outperforming several SOTA CNNs. 
The result is promising, and the model has great potential to assist ophthalmologists. 
The code for implementing this work is available from \url{https://github.com/gyanendar/InceptionCaps}.

The completion of this work was fraught with difficulties. Several research papers that describe automatic glaucoma classification from fundus images use private datasets and omit key information about the model architecture. As a result, those studies' findings cannot be independently validated. The results of four separate SOTA models were replicated using publicly available datasets to ensure their applicability.

Furthermore, no standard benchmark indicators are employed to assess the model performance uniformly, making the comparison of the models impossible. Consequently, multiple performance metrics were utilised in order to compare and contrast model performance.

Future work will include training the InceptionV3 model on a large retinal image dataset for InceptionCaps and should be tested on a diverse
dataset to demonstrate its usefulness for glaucoma classification. The CNNs are complex models and increase the computational resource necessary. More work is needed to optimize the models' computational resource requirement by compressing the model through pruning and quantization.

\end{document}